%% file: main.tex
\documentclass[conference]{IEEEtran}
\IEEEoverridecommandlockouts
\usepackage{cite}
\def\BibTeX{{\rm B\kern-.05em{\sc i\kern-.025em b}\kern-.08em
    T\kern-.1667em\lower.7ex\hbox{E}\kern-.125emX}}

\usepackage[utf8]{inputenc} 
\usepackage[T1]{fontenc}    
\usepackage{hyperref}       
\usepackage{url}            
\usepackage{booktabs}       
\usepackage{amsfonts}       
\usepackage{nicefrac}       
\usepackage{microtype}      

\usepackage{todonotes}
\usepackage{listings}
\usepackage[cmex10]{amsmath}
\usepackage{amssymb}
\usepackage[linesnumbered,ruled,vlined]{algorithm2e}
\definecolor{cmt_green}{RGB}{0, 128, 0}

\SetCommentSty{mycommfont}
\usepackage{bm}

\usepackage{array}
\usepackage{mdwmath}
\usepackage{mdwtab}
\usepackage{dirtree}
\usepackage{subfig}
\usepackage{multirow}
\usepackage{multicol}
\usepackage{svg}

\usepackage{mathtools}

\definecolor{Maroon}{rgb}{0.5,0,0}
\definecolor{DarkOliveGreen}{rgb}{0,0.5,0}

\definecolor{maroon}{cmyk}{0, 0.87, 0.68, 0.32}
\definecolor{halfgray}{gray}{0.55}
\definecolor{ipython_frame}{RGB}{207, 207, 207}
\definecolor{ipython_bg}{RGB}{247, 247, 247}
\definecolor{ipython_red}{RGB}{186, 33, 33}
\definecolor{ipython_green}{RGB}{0, 128, 0}
\definecolor{ipython_cyan}{RGB}{64, 128, 128}
\definecolor{ipython_purple}{RGB}{170, 34, 255}

\lstdefinelanguage{XML}
{
	basicstyle=\scriptsize,
	morestring=[b]",
	moredelim=[s][\bfseries\color{Maroon}]{<}{\ },
	moredelim=[s][\bfseries\color{Maroon}]{</}{>},
	moredelim=[l][\bfseries\color{Maroon}]{/>},
	moredelim=[l][\bfseries\color{Maroon}]{>},
	morecomment=[s]{<?}{?>},
	morecomment=[s]{<!--}{-->},
	commentstyle=\color{DarkOliveGreen},
	stringstyle=\color{blue},
	identifierstyle=\color{red},
	rulecolor=\color{ipython_frame},
	frame=single,
	backgroundcolor=\color{ipython_bg},
	tabsize=2,
}

\lstdefinelanguage{python}{
	morekeywords={access,and,break,class,continue,def,del,elif,else,except,exec,finally,for,from,global,if,import,in,is,lambda,not,or,pass,print,raise,return,try,while},
	morekeywords=[2]{abs,all,any,basestring,bin,bool,bytearray,callable,chr,classmethod,cmp,compile,complex,delattr,dict,dir,divmod,enumerate,eval,execfile,file,filter,float,format,frozenset,getattr,globals,hasattr,hash,help,hex,id,input,int,isinstance,issubclass,iter,len,list,locals,long,map,max,memoryview,min,next,object,oct,open,ord,pow,property,range,raw_input,reduce,reload,repr,reversed,round,set,setattr,slice,sorted,staticmethod,str,sum,super,tuple,type,unichr,unicode,vars,xrange,zip,apply,buffer,coerce,intern},
	sensitive=true,
	morecomment=[l]\#,
	morestring=[b]',
	morestring=[b]",
	morestring=[s]{'''}{'''},
	morestring=[s]{"""}{"""},
	morestring=[s]{r'}{'},
	morestring=[s]{r"}{"},
	morestring=[s]{r'''}{'''},
	morestring=[s]{r"""}{"""},
	morestring=[s]{u'}{'},
	morestring=[s]{u"}{"},
	morestring=[s]{u'''}{'''},
	morestring=[s]{u"""}{"""},
	literate=
	{á}{{\'a}}1 {é}{{\'e}}1 {í}{{\'i}}1 {ó}{{\'o}}1 {ú}{{\'u}}1
	{Á}{{\'A}}1 {É}{{\'E}}1 {Í}{{\'I}}1 {Ó}{{\'O}}1 {Ú}{{\'U}}1
	{à}{{\`a}}1 {è}{{\`e}}1 {ì}{{\`i}}1 {ò}{{\`o}}1 {ù}{{\`u}}1
	{À}{{\`A}}1 {È}{{\'E}}1 {Ì}{{\`I}}1 {Ò}{{\`O}}1 {Ù}{{\`U}}1
	{ä}{{\"a}}1 {ë}{{\"e}}1 {ï}{{\"i}}1 {ö}{{\"o}}1 {ü}{{\"u}}1
	{Ä}{{\"A}}1 {Ë}{{\"E}}1 {Ï}{{\"I}}1 {Ö}{{\"O}}1 {Ü}{{\"U}}1
	{â}{{\^a}}1 {ê}{{\^e}}1 {î}{{\^i}}1 {ô}{{\^o}}1 {û}{{\^u}}1
	{Â}{{\^A}}1 {Ê}{{\^E}}1 {Î}{{\^I}}1 {Ô}{{\^O}}1 {Û}{{\^U}}1
	{œ}{{\oe}}1 {Œ}{{\OE}}1 {æ}{{\ae}}1 {Æ}{{\AE}}1 {ß}{{\ss}}1
	{ç}{{\c c}}1 {Ç}{{\c C}}1 {ø}{{\o}}1 {å}{{\r a}}1 {Å}{{\r A}}1
	{€}{{\EUR}}1 {£}{{\pounds}}1
	{^}{{{\color{ipython_purple}\^{}}}}1
	{=}{{{\color{ipython_purple}=}}}1
	{+}{{{\color{ipython_purple}+}}}1
	{*}{{{\color{ipython_purple}$^\ast$}}}1
	{/}{{{\color{ipython_purple}/}}}1
	{+=}{{{+=}}}1
	{-=}{{{-=}}}1
	{*=}{{{$^\ast$=}}}1
	{/=}{{{/=}}}1,
	literate=
	*{-}{{{\color{ipython_purple}-}}}1
	{?}{{{\color{ipython_purple}?}}}1,
	identifierstyle=\color{black}\ttfamily,
	commentstyle=\color{ipython_cyan}\ttfamily,
	stringstyle=\color{ipython_red}\ttfamily,
	keepspaces=true,
	showspaces=false,
	showstringspaces=false,
	rulecolor=\color{ipython_frame},
	frame=single,
	backgroundcolor=\color{ipython_bg},
	basicstyle=\scriptsize,
	keywordstyle=\color{ipython_green}\ttfamily,
}

\usepackage{mathtools}
\usepackage{cuted}

\begin{document}

\title{Ensemble Transfer Learning for Emergency Landing Field Identification on Moderate Resource Heterogeneous Kubernetes Cluster}

\author{\IEEEauthorblockN{1\textsuperscript{st} Andreas Klos}
\IEEEauthorblockA{\textit{Chair of computer architecture} \\
\textit{FernUniversität in Hagen}\\
Hagen, Germany \\
andreas.klos@fernuni-hagen.de}
\and
\IEEEauthorblockN{2\textsuperscript{nd} Marius Rosenbaum}
\IEEEauthorblockA{\textit{Chair of computer architecture} \\
\textit{FernUniversität in Hagen}\\
Hagen, Germany \\
marius.rosenbaum@fernuni-hagen.de}
\and
\IEEEauthorblockN{3\textsuperscript{rd} Wolfram Schiffmann}
\IEEEauthorblockA{\textit{Chair of computer architecture} \\
\textit{FernUniversität in Hagen}\\
Hagen, Germany \\
wolfram.schiffmann@fernuni-hagen.de}
}


\maketitle

\input{abstract}
\begin{IEEEkeywords}
Transfer learning, Ensemble learning, Kubernetes, Emergency landing field, Bandit optimization, Bayesian optimization
\end{IEEEkeywords}

\input{chapter1/chapter1}

\input{chapter2/chapter2}

\input{chapter3/chapter3}

\input{chapter4/chapter4}

\section*{Acknowledgment}
The authors want to thank Mr. Florian Fromm for his recommendations regarding \texttt{Kubernetes}, \texttt{RabbitMQ}, \texttt{Prometheus} and \texttt{Grafana} as well as for proof reading the corresponding subsection.

\bibliographystyle{ieeetr}

\bibliography{lit.bib}

\end{document}

%% file: abstract.tex
\begin{abstract}
    The full loss of thrust of an aircraft requires fast and reliable decisions of the pilot. If no published landing field is within reach, an emergency landing field must be selected. The choice of a suitable emergency landing field denotes a crucial task to avoid unnecessary damage of the aircraft, risk for the civil population as well as the crew and all passengers on board. Especially in case of instrument meteorological conditions it is indispensable to use a database of suitable emergency landing fields. Thus, based on public available digital orthographic photos and digital surface models, we created various datasets with different sample sizes to facilitate training and testing of neural networks. Each dataset consists of a set of data layers. The best compositions of these data layers as well as the best performing transfer learning models are selected. Subsequently, certain hyperparameters of the chosen models for each sample size are optimized with Bayesian and Bandit optimization. The hyperparameter tuning is performed with a self-made Kubernetes cluster. The models outputs were investigated with respect to the input data by the utilization of layer-wise relevance propagation. With optimized models we created an ensemble model to improve the segmentation performance. Finally, an area around the airport of Arnsberg in North Rhine-Westphalia was segmented and emergency landing fields are identified, while the verification of the final approach's obstacle clearance is left unconsidered. These emergency landing fields are stored in a PostgreSQL database. 
\end{abstract}

%% file: chapter1/chapter1.tex
\section{Introduction}

The loss of thrust depicts a major issue for every pilot. In such a stressful situation quick and focused action is required. There are several reasons why the engines of an aircraft can fail completely. For example, a bird strike on all engines such as in case of flight UA1549 in 2009 and the subsequent forced landing in the Hudson or a technical problem on the single engine of a general aviation aircraft. In those emergency cases each aircraft becomes a glider and the pilot must choose a suitable glide path so that the aircraft arrives at an appropriate altitude at the beginning of the selected landing field e.\,g. as described in \cite{ShMa2017,MKlein2018}. Thereby, the reachability of landing fields is limited by the residual altitude of the aircraft at the time of incident.

Furthermore, it is not guaranteed that an published landing field -- in the best case paved -- is within reach. In that case, the pilot is compelled to select an emergency landing field which is mainly based on his experience and the emergency guidelines. During the choice of a suitable emergency landing field several terrain properties like the size, shape, slope, surface, surrounding and civilization as well as the current conditions e.\,g. the wind, season of the year, rainfall etc. have to be considered. 

The selection of an appropriate emergency landing field is a crucial task and influences the degree of possible damage of the aircraft and viability of the crew members as well as the passengers. For that reason, our objective is the acceleration of the pilots decision process by providing a database with appropriate emergency landing fields for the specific aircraft type. 

For the autonomous identification of emergency landing fields (ELFs), many machine vision and machine learning techniques have been proposed in recent years which can be subdivided in three categories: 1) processing real-time images obtained by on-board sensors; 2) processing pre-acquired data; 3) processing multi-modal images sources.

In \cite{MCLP+2019} an embeddable Convolutional Neural Networks (CNNs) trained on synthetic data to estimate the obstacle clearness and safeness of regions for landing an UAV is proposed. The approach is tested with UAV footage. Every pixel is assigned to one of the categories: horizontal (landable), vertical (obstacles) and others (safer for a landing). Unfortunately, the classification of flat areas as suitable for an landing leads also to classifications of highways and water aerials as appropriate. In \cite{RAKB+2016} an $k$-Nearest Neighbor approach that considers a feature vector of data acquired by an UAV camera and measures of a light intensity sensor is proposed. Regrettably, only small areas within the field of view are analyzed depending on meteorological conditions. Thus, only a highly restricted number of suitable ELFs can be found. These drawbacks are revised in \cite{SRKL2013} by aircraft-mounted cameras oriented to the front and a horizon detection algorithm to identify the ground in the images. Besides, they apply a nonlinear retinex image-enhancement method to revamp the environmental effects and improve the contrast and sharpness. The results depending on the resolution of the aircraft-mounted cameras and the altitude of the aircraft. In \cite{Mej2014}  a surface classification of ELFs is introduced. The classification is performed by a multi-class Support Vector Machine (SVM). Other terrain classifications are proposed in \cite{NaPe2012} -- SVM and AdaBoost for multi-spectral images, in \cite{KSL2010} -- SVM and multi layer perceptron, and in \cite{CKJ2010} -- premised on SVM and Random Forrests processing monocular camera data. In \cite{FWC2005} an algorithm is introduced which applies standard image processing techniques and artificial neural networks to verify obstacle clearness. 

In \cite{LOP2014} a two-stage segmentation approach based on satellite imagery is proposed. First, an initial segmentation is performed by analyzing the corresponding histogram of the satellite imagery to estimate the number of different classes. Afterwards, a structure preserving segmentation is performed in the spectral domain. This approach lacks in its ability of edge detection and disregards the suitability for an emergency landing due to its length and width. In \cite{MeFi2013} another two step processing algorithm is presented which first performs a sectioning of the considered region -- Canny edge, line growing -- and subsequently a geometric check to ensure the suitability regarding to the shape as well as dimension of the examined region. These image processing steps omit to analyze the slope and bumpiness which are required to guarantee the suitability of located ELFs. In \cite{GKS2015} a digital elevation model processing approach is introduced which performs the examination by the quadtree data structure. The metric of variance and average altitude may be insufficient for the selection of a suitable ELF because outliers -- caused e.\,g. by buildings -- might be pruned which could lead to a false-negative classification of the corresponding region. Besides, the lack of investigation of the ELF's surface could hide e.\,g. water areas which depicts also a weakness to the algorithm proposed in \cite{EWS2018} where only elevation data is processed regarding to predefined slope restrictions. In \cite{FUY2018} a patch based segmentation approach is proposed based on a CNN aerial images acquired from \texttt{Google Maps}. The classification is performed into the following three categories: Safe, Not recommended, Other. The CNNs performance is evaluated by considering the precision scores achieved for the safe areas (63.8\%) and for both, the safe and the not recommended regions (87.1\%). The exclusive processing of satellite imagery neglects the bumpiness and might result in false negative classification regarding the landability. A semi-automated emergency landing site selection algorithm is proposed in \cite{ALUB+2018} operating on \texttt{Google Maps} sattelite imagery, digital elevation models and a human settlement layer. The segmentation is based on standard image processing methods. Besides, the safety estimation of the ELF consideres five different measures. Furthermore, a reachability analysis is performed.

Hence, a combination of processing images from an aircraft-mounted camera and pre-acquired DEMs is shown in \cite{WMYA+2015}. The authors investigated the processing of 2D geodata and reconstructed 3D model. In \cite{MFEL2009} another multi-modal processing algorithm is proposed. First, preliminary processing steps are performed as mentioned earlier in \cite{MeFr2013}. Afterwards, man-made and natural objects were distinguished by considering the intensity values. Subsequently, the geometric shape, the surface type and the slope are considered. Unfortunately, the obstacle clearance of the final approach is left unconsidered.

In this paper, we present a patch segmentation of multi-modal geodata based on pre-trained artificial neural networks (ANNs) by the application of ensemble learning. Our implementations are mainly realized in \texttt{Python} with the usage of the \texttt{PyTorch API} and a self-made heterogeneous moderate resource \texttt{Kubernetes} cluster. The manual segmentation of the regions has been evaluated by thresholding so that false-negative classifications are avoided or at least reduced. The datasets are manual labeled by the utilization of \texttt{QGIS}. Our approach utilizes data fusion of the digital surface model and orthophotos to train, validate and test the selected ANNs. After training the selected ANNs, the ensemble model is created and applied to an area around the airport of Arnsberg, in North Rhine-Westphalia. Subsequently, the areas identified as suitable for an emergency landing by the patch segmentation are stored in a \texttt{PostgreSQL} database as georeferenced polygons. Further geographic queries are performed by the usage of the extension \texttt{PostGIS} -- a spatial database extender of \texttt{PostgreSQL} -- to identify runways. These runways are also stored in the database. In the prior mentioned area we identified 54,997 ELFs and were able to segment 26.252\,m$^2$ as suitable for an emergency landing and 221.329\,m$^2$ as unlandable.

The paper is structured as follows: In Sec. 2 the utilized dataset, its generation as well as the constructed deeplearning infrastructure are proposed. Afterwards, in Sec. 3 our investigations are proposed and the achieved results are presented as well as discussed. In Sec. 4 the paper finalizes with a conclusion and an outlook on our future works.

%% file: chapter2/chapter2.tex
\section{Dataset generation and deeplearning infrastructure}

\subsection{Dataset generation}

To the knowledge of the authors, due to the time of our research, no public available dataset did exist for the training, validation and test of ANNs regarding classification of multi-modal geodata as landable or unlandable. Therefore, we created different datasets for supervised learning.

\begin{figure}[ht]
\centering
    \subfloat[RGB]{
        \includegraphics[width=.15\textwidth]{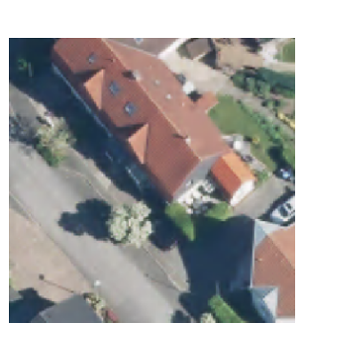}
    }
    \subfloat[NIR]{
        \includegraphics[width=.15\textwidth]{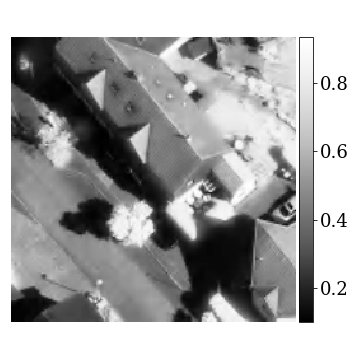}
    }
    \subfloat[NDVI]{
        \includegraphics[width=.155\textwidth]{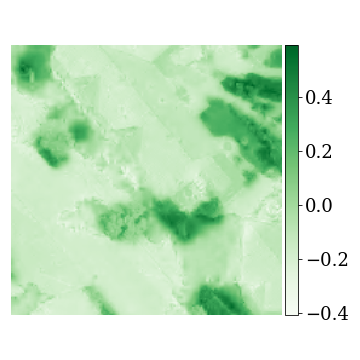}
    }\\
    \subfloat[Inter. DSM]{
        \includegraphics[width=.151\textwidth]{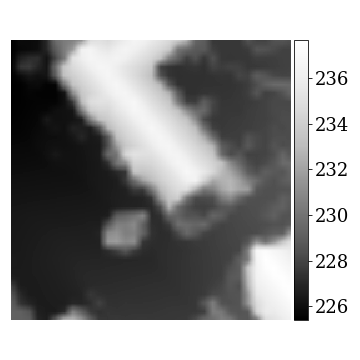}
    }
    \subfloat[Roughness]{
        \includegraphics[width=.146\textwidth]{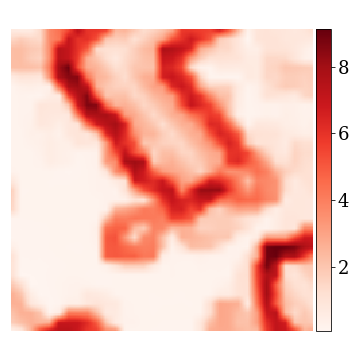}
    }
    \subfloat[Slope]{
        \includegraphics[width=.15\textwidth]{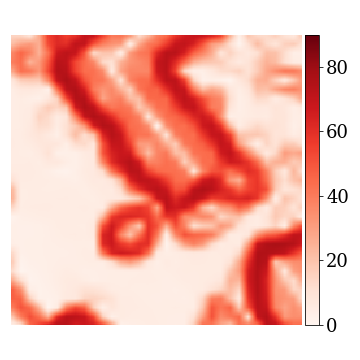}
    }
    \caption{Decomposed unlandable input sample.}
    \label{fig:c3:in}
\end{figure}

The unlabeled and raw geodata -- spatial reference system: EPSG 25832 (ETRS89 / UTM Zone 32N) -- is downloaded from \cite{dom} and \cite{dop}. This data is composed of digital orthographic photos (DOP) with four channels (red, green, blue, near infrared with horizontal resolution of 0.2\,m per pixel as shown in Fig. \ref{fig:c3:in} (a) and (b)) and digital surface models (DSM) (point clouds with X, Y position and altitude with horizontal resolution of four points m$^2$ and 0.2\,m in vertical direction).

The point clouds are interpolated by inverse distance weighting with the \texttt{gdal} Python package configured as follows: \texttt{invdistnn:power=2.0}, \texttt{radius=1.415}, \texttt{max\_points=16}, \texttt{nodata=-2147483648.0}. The resulting resolution of the interpolated raster is 1\,m per pixel and 0.2\,m in altitude. The interpolated DSMs and the DOPs were stored in a \texttt{PostgreSQL} database with the \texttt{PostGIS} extension.

Subsequently, the data became labeled by quadratic polygons with various sizes (32 m$^2$, 64 m$^2$, 128 m$^2$, 256 m$^2$). The polygons are labeled with 0 and 1. Thereby, 0 denotes unlandable and 1 landable areas. Afterwards, the individual examples are queried from the database with respect to the corresponding polygons. Each polygon determines an area for which the corresponding DOP and DSM are queried from the database. The elevation data is interpolated to the same horizontal resolution as the DOP by the bilinear interpolation algorithm \cite[P. 88]{Gonzalez}. Additionally, the roughness and slope is calculated by the usage of \texttt{PostGIS} standard functions (\texttt{ST\_Roughness}, \texttt{ST\_Slope}) as illustrated in Fig. \ref{fig:c3:in} (e) and (f). The color ranges from white to red where the darkness of red determines the elevation difference and the slope. Besides, the normalized difference vegetation index (NDVI) is computed: ($(\text{NIR}-\text{Red})/(\text{NIR}+\text{Red})$). An NDVI image is shown in Fig. \ref{fig:c3:in} (c) where the color ranges from white to green. The hue of the green color determines the biomass at the considered position.

Each queried area is subsampled in different search windows (SW) of 8\,m$^2$, 16\,m$^2$ and 32\,m$^2$ and stride $\frac{\text{SW}}{2}$. Furthermore, each example tagged as landable is evaluated regarding their slope by thresholding. The results are stored in a *.hdf5 file. The sample count of the various generated dataset is as follows: SW 8 m$^2$: \{train: 380,928 with \{0: 190,464, 1: 190,464\}, test: 76,288 with \{0: 38,152, 1: 38,136\}\}, SW 16 m$^2$: \{train: 84,992 with \{0: 42,498, 1: 42,494\}, test: 17,024 with \{0: 8,516, 1: 8,508\}\}, SW 32 m$^2$: \{train: 16,768 with \{0: 8,424, 1: 8,344\}, test: 3,328 with \{0: 1,672, 1: 1,656\}\}.

\subsection{Deeplearning infrastructure}

The subsequent results are achieved by the utilization of a heterogeneous cluster with high requirements regarding availability and reliability. Therefore, we build a cluster based on \texttt{Kubernetes}\footnote{For further information the authors refer to \cite{Luk2018}} (k8s) with the \texttt{Docker} engine and the \texttt{NVIDIA Container Toolkit}. To facilitate the usage of GPUs, the default runtime of \texttt{Docker} is changed to \texttt{NVIDIA}. The nodes are composed of a master, several workers and a network file system (NFS) server. The Master is depicted by an personal computer (PC) dedicated for constantly serving the cluster. The PCs of the chair's employees as well as one PC provided for always serving the cluster -- further denoted as Worker 1 -- are the workers. Every worker is configured with an dual boot operating system (OS). If the employee decides, that the PC is currently free for doing some other task, the PC can be booted as Ubuntu 18.04 LTS OS and the worker will be automatically tagged as ready in the k8s cluster. Subsequently, a \texttt{Pod} is launched on the new, ready node. The NFS-Server is launched on Worker 1 to provide the NFS as volume mount on each worker. The hardware configuration of the k8s cluster is detailed in Tab. \ref{tab:hw}.

\begin{table}[ht]\caption{Hardware configuration of the k8s cluster.}
    \centering\resizebox{.49\textwidth}{!}{%
    \begin{tabular}{l|c|c|c}\hline
        \textbf{Node}&\textbf{NVIDIA GeForce GPU}&\textbf{CPU}&\textbf{RAM}  \\\hline
        Master&\texttt{GTX 1080 \hspace{5cm}}&\texttt{i7-2600K}&12\,GiB DDR3\\\hline
        Worker 1&\texttt{GTX 1080 Ti}&\texttt{i7-6700K}&32\,GiB DDR4\\\hline
        Worker 2&\texttt{RTX 2080 Ti}&\texttt{i9-9900K}&32\,GiB DDR4\\\hline
        Worker 3&\texttt{GTX 1080 Ti}&\texttt{i7-7700K}&32\,GiB DDR4\\\hline
        Worker 4&\texttt{GTX 1080}&\texttt{i7-7700K}&32\,GiB DDR4\\\hline
    \end{tabular}}
    \label{tab:hw}
\end{table}


Additionally, a \texttt{RabbitMQ} message broker is deployed as \texttt{StatefulSet}. The message broker is used for delivering tasks from a queue to its consuming clients. Inside the k8s cluster, the message broker becomes accessible by a dedicated service. This service guarantees the reachability through the assigned DNS entry and the required ports at anytime. The message broker exploits a persistent volume claim to request a persistent volume with \texttt{ReadWriteOne} access mode and a capacity of 1\,GiB. To facilitate a high data reliability, the persistent volume is mounted at \texttt{RabbitMQ}'s data storage location in \texttt{/var/lib/rabbitmq}. The message broker is used for distributing the tasks to each worker node. The message broker is configured as follows: 1) Manual message acknowledgments to make sure that a message (task) is never lost e.\,g. caused by a closed connection, 2) Durable queues and persistent messages which ensures the survival of the queues as well as the messages even if \texttt{RabbitMQ} restarts, 3) Fair message dispatching by setting the prefetch count equal one which configures \texttt{RabbitMQ} to give only one unacknowledged message to a worker at a time.

Furthermore, a \texttt{PostgreSQL} database server is deployed as a \texttt{StatefulSet}. A connection to the database can be established by the usage of the corresponding service as described before. The database uses a persistent volume claim to request a persistent volume. To provide high data reliability, the persistent volume is mounted at the storage location of the database at \texttt{/var/lib/postgresql/data} which guarantees a high reliability of the data. The purpose of the database is the recording of the current state of each worker regarding the task in process accompanied by the storage of the best result achieved for each task during validation and test phase.

The message broker and database are assigned to Worker 1. Other relevant deployments launch their \texttt{Pod}s on those node, e.\,g. \texttt{CoreDNS}, otherwise the shutdown of one of the other workers (Worker $\text{n with n} \in {2,...,4}$) might result in an error.

Besides, a deployment is created which provides declarative updates for \texttt{Pod}s and the corresponding \texttt{ReplicaSet}. The \texttt{ReplicaSet} launches a \texttt{Pod} on each worker. Each process -- a containerized task, further called worker-process -- running in the container inside the \texttt{Pod} subscribes the message broker. As soon as a queue with tasks is created in the message broker, an individual task is send to each worker-process. The next task will be first assigned to one of the listening worker-processes, if a successful completion confirmation is send to the message broker. The results achieved by the worker-process are stored in the aforementioned database. Each task is depicted by a string with the following information: \texttt{"searchWindow:dataComposition:modelName"}. The last task in the queue triggers an automatic selection of the task, with the best test results regarding the accuracy for each search size. Subsequently a new task queue is created and send to the message broker. After receiving the new queue, the novel tasks are distributed as before mentioned to each worker-process by the message broker.

The monitoring of the k8s cluster is performed with the \texttt{Helm} charts \texttt{Prometheus} and \texttt{Grafana}. Periodic backups are done with \texttt{Velero} and a \texttt{Minio S3 Object Store}.

\subsection{Emergency landing field dimension}

The necessary width of the ELF is determined by the wing span of the airplane. In our case, we consider the DA20 C1  with an wing span of 10.89\,m \cite{DA20}. For safety purposes, we increase the necessary width of the ELF by the factor three (32.67\,m). The required ELF length -- rolling distance -- can be estimated according to \cite{Ray2018} and \cite{Coo2016}. The equations have been adjusted to incorporate a possible inclination of the ELF and the free ground roll distance without hitting the breaks. Equation \ref{eq:acc} describes how the acceleration during landing -- ground roll -- can be calculated.

\begin{equation}\label{eq:acc}
    a = \frac{g}{W}\cdot[T-W_x-D-\mu\cdot R]
\end{equation}

Thereby, $g$ denotes the gravitational acceleration, $W$ is the gravitational force, $T$ is the force caused by thrust, $W_x$ is the downhill force, $D$ depicts the drag force, $\mu$ is the rolling friction coefficient and R the weight force on the wheels ($W_y-L$ with L as the lift drag). In Eq. \ref{eq:acc_full} the calculation of the acceleration is further decomposed to present the necessary variables to calculate.
\begin{equation}\label{eq:acc_full}
    a = \frac{g}{W}\cdot[T-W\cdot \sin(\alpha)-D-\mu\cdot (W\cdot \cos(\alpha)-L)]
\end{equation}
where $\alpha$ depicts the inclination angle of the ELF. The computation of the lift force is presented in Eq. \ref{eq:lift}
\begin{equation}\label{eq:lift}
    L = q\cdot S\cdot C_L
\end{equation}
where $q$ is the dynamic pressure, $S$ depicts the wing area and $C_L$ is the lift coefficient. In Eq. \ref{eq:dynPressure} is formalized, how the dynamic pressure is calculated.

\begin{equation}\label{eq:dynPressure}
    q = \frac{1}{2} \cdot \rho \cdot V^2
\end{equation}

The $\rho$ is the is the air density and $V$ the velocity of the aircraft in the surround medium.  In Eq. \ref{eq:drag} is the calculation of the drag force shown
\begin{equation}\label{eq:drag}
    D = q\cdot S\cdot C_D
\end{equation}

where the $C_D$ is the drag coefficient. Under the assumption of an uncambered wing profile, the drag coefficient can be expressed by Eq. \ref{eq:dragCo}.
\begin{equation}\label{eq:dragCo}
    C_D = C_{D_0} + K\cdot C_L^2
\end{equation}

Thereby, $C_{D_0}$ denotes the zero-lift drag coefficient, $K\cdot C_L^2$ is the induced drag coefficient. The formalism of zero-lift drag coefficient calculatiuon is shown in Eq. \ref{eq:zeroliftDrag}.

\begin{equation}\label{eq:zeroliftDrag}
    C_{D_0} = \frac{1}{(2\cdot L/D_{max})^2\cdot K}
\end{equation}

Due to the assumption of zero thrust during the cosnidered emergency situation, $L/D_{max}$ equals the Glide number $\left(\frac{1}{tan(\alpha)} \text{ with alpha as angle of attack}\right)$. The drag due to lift factor $K$ is computed as represented by Eq. \ref{eq:K}.

\begin{equation}\label{eq:K}
    K = \frac{1}{\pi\cdot A \cdot e}
\end{equation}

Obviously, the drag due to lift factor depends on the aspect ration $A$ and Oswald's span efficiency factor $e$. The calculation of $A$ is shown in Eq. \ref{eq:aspectratio}
\begin{equation}\label{eq:aspectratio}
    A = \frac{b^2}{S}
\end{equation}
where $b$ is the wing span. Equation \ref{eq:oswald} presents an estimation of $e$.
\begin{equation}\label{eq:oswald}
    e = 1.78\cdot (1-0.045\cdot A^{0.68})-0.64
\end{equation}
Typical values for $e$ range from 0.7 to 0.85. In Eq. \ref{eq:liftCo} is shown, how the lift coefficient can be computed under the assumption, that $L$ equals $W$.

\begin{equation}\label{eq:liftCo}
    C_L = \frac{W}{q\cdot S}
\end{equation}

With Eq. 2 - 11 in mind, the acceleration can be calculated as presented by Eq. \ref{eq:acc3}.
\begin{equation}\label{eq:acc3}
    a = g\cdot (K_A+K_T\cdot V^2)
\end{equation}

During landing operation, $K_T$ and $K_A$ are assumed as constants. Thereby, $K_T$ includes the thrust related and $K_A$ the aerodynamic related terms as in Eq. \ref{eq:ktka}.

\begin{equation}\label{eq:ktka}
    \begin{split}
        K_T = \frac{T}{W} - \sin(\alpha)-\mu \cdot \cos(\alpha)\\
        K_A = \frac{\rho\cdot S}{2\cdot W}\cdot (\mu C_L - C_{D_0} - K\cdot C_L^2)
    \end{split}
\end{equation}

Equation \ref{eq:sg1} shows, how the necessary ground roll distance can be computed, by integrating the expression $\frac{V}{a}$ from the touch down velocity $V_{td}$ and the final velocity $V_f$.

\begin{equation}\label{eq:sg1}
    s_g = s_{fr}+\int_{V_{td}}^{V_f} \frac{V}{a}\cdot dV
\end{equation}

The $s_{fr}$ denotes the free roll distance without hitting the breaks and is calculated by the touch down velocity times the reaction time for hitting the breaks -- assumed as 3\,s. The aforementioned integration results in Eq. \ref{eq:sg2}.

\begin{equation}\label{eq:sg2}
    s_g = s_{fr}+\frac{1}{2\cdot g\cdot K_A}\cdot ln\left(\frac{K_T+K_A\cdot V_f^2}{K_T+K_A\cdot V_{td}^2}\right)
\end{equation}

Due to the fact that the $V_f$ equals zero, the equation can be further simplified as shown in Eq. \ref{eq:sg3}.

\begin{equation}\label{eq:sg3}
    s_g = s_{fr}+\frac{1}{2\cdot g\cdot K_A}\cdot ln\left(\frac{K_T}{K_T+K_A\cdot V_{td}^2}\right)
\end{equation}

Under the assumption of zero thrust during the emergency procedure, $K_T$ can be reduced to Eq. \ref{eq:kt}.

\begin{equation}\label{eq:kt}
    K_T = K_{T_{0}} =  -\sin(\alpha)-\mu \cdot \cos(\alpha)
\end{equation}

Inserting the constants $K_T$ and $K_A$ into Eq. \ref{eq:sg3} results in Eq. \ref{eq:sgfinal}.
\begin{figure*}[bt]
    \centering
        \begin{equation}\label{eq:sgfinal}
            s_g = V_{td}\cdot t_{r}+\frac{1}{2\cdot g\cdot \frac{\rho\cdot S}{2\cdot W}\cdot (\mu C_L - C_{D_0} - K\cdot C_L^2)}\cdot ln\left(\frac{-\sin(\alpha)-\mu \cdot \cos(\alpha)}{-\sin(\alpha)-\mu \cdot \cos(\alpha)+\frac{\rho\cdot S}{2\cdot W}\cdot (\mu C_L - C_{D_0} - K\cdot C_L^2)\cdot V_{td}^2}\right)
        \end{equation}    
\end{figure*}

The following assumptions are done, for calculating the minimum required ELF length which sums up to 210.773\,m: $m$ = 800\,kg, temperature  = 15$^\circ$C, $\rho$ = 1.225\,$\frac{kg}{m^3}$, $g$ = 9.807\,$\frac{m}{s^2}$, $\alpha$ = 0$^\circ$, $S$ = 11.6\,m$^2$, $b$ = 10.89\,m, $\mu$ = 0.2 (soft turf and brakes on), $t_r$ = 3\,s, $V_{td}$ = $1.15\cdot V_{stall}$ = 21.298\,$\frac{m}{s}$ (stall speed during landing with 0$^\circ$ bank angle), $L/D_{max}$ = 11. The European Aviation Safety Agency (EASA) recommends for the surface type grass (on firm soil up to 20\,cm long) in \cite{EASA2014} to increase the runway length by the factor 1.15 which results in 242.389\,m.

Most ELFs can be used in two landing directions. The selection of the preferable direction depends on the wind force and direction, the reachability as well as the obstacle clearance of the final approach. This considerations are omitted in following. For descending ELFs the EASA recommends to extend the minimum required ELF length by 5\% per 1\% of downslope. 
 
The slope angle is computed by considering the elevation points covered by the center line of each identified ELF. In the resulting point cloud, a linear regression is performed and a line is fitted into the point cloud. Furthermore, the difference between the elevation values at the start and end point is calculated. The maximum absolute slope of both approaches is utilized to determine $\alpha$ in Eq. \ref{eq:acc} - \ref{eq:sgfinal}. If the $\alpha$ is negative, the recommendation of EASA is followed and per percent of downslope the minimum required ELF length becomes increased by 5\%.

%% file: chapter3/chapter3.tex
\section{Results and discussion}

\subsection{Best input data  composition}\label{subsec:dc}
It is assumed, that each search size benefits from a certain input data composition. For proofing the assumption, we made a analysis of the following input data compositions on the basis of AlexNet\cite{KSH2012}: RGB, NIR, SLOPE, ROUGHNESS, NDVI, DOM, RGB-NIR, RGB-Slope, RGB-NDVI, NIR-Slope, NDVI-Slope, NDVI-NIR, RGB-NIR-Slope, NDVI-NIR-Slope, RGB-NIR-NDVI-Slope. All models are trained four times, maximum 100 epochs, with batch size 128\footnote{If an out of memory (OOM) exception is catched, the batch size becomes halved till the training of the model fits into memory of the GPU. } and early stopping, after 6 Epochs of no improvement. The mean squared error (MSE) loss function and the Adam optimizer\cite{KiBa2014} are chosen. The parameterization of the Adam optimizer is equal to the default settings of \texttt{PyTorch} framework except weight decay ($10^{-5}$). No data augmentation is applied to the input data so that the evaluation of the best data composition is more meaningful. The best results achieved regarding accuracy on test dataset are shown in Tab. \ref{tab:best_res_data}.

\begin{table}[ht]\caption{The results achieved for training, validation and test of AlexNet on the corresponding input data composition. The abbreviation SW denotes the search size and T the duration time.}
    \centering\resizebox{.49\textwidth}{!}{%
    \begin{tabular}{c|c|c|c|c|c|c|c|c}
        \hline
        \multirow{1}{*}{\textbf{SW}} & \multirow{1}{*}{\textbf{Data}} & \multicolumn{3}{c|}{\textbf{Accuracy [\%]}} & \multicolumn{3}{c|}{\textbf{Loss $\pmb{[10^{-3}]}$}} & \multirow{1}{*}{\textbf{T}}\\
        $\pmb{[m^2]}$&\textbf{composition}&\textbf{train}&\textbf{valid}&\textbf{test}&\textbf{train}&\textbf{valid}&\textbf{test}&\textbf{[s]}\\\hline
        8& RGB-NIR-Slope &99.398&99.603&99.565&9.185&6.811&7.027&552\\\hline
        16& NDVI-Slope &99.728&99.811&99.666&6.163&4.433&5.420&54\\\hline
        32& RGB-Slope &99.854&99.917&99.970&5.735&5.886&5.644&19\\\hline
    \end{tabular}}
    \label{tab:best_res_data}
\end{table}

\begin{figure*}[ht]
    \centering
    \subfloat[Data compositions, SW 8 m$^2$]{
        \includegraphics[width=.32\textwidth]{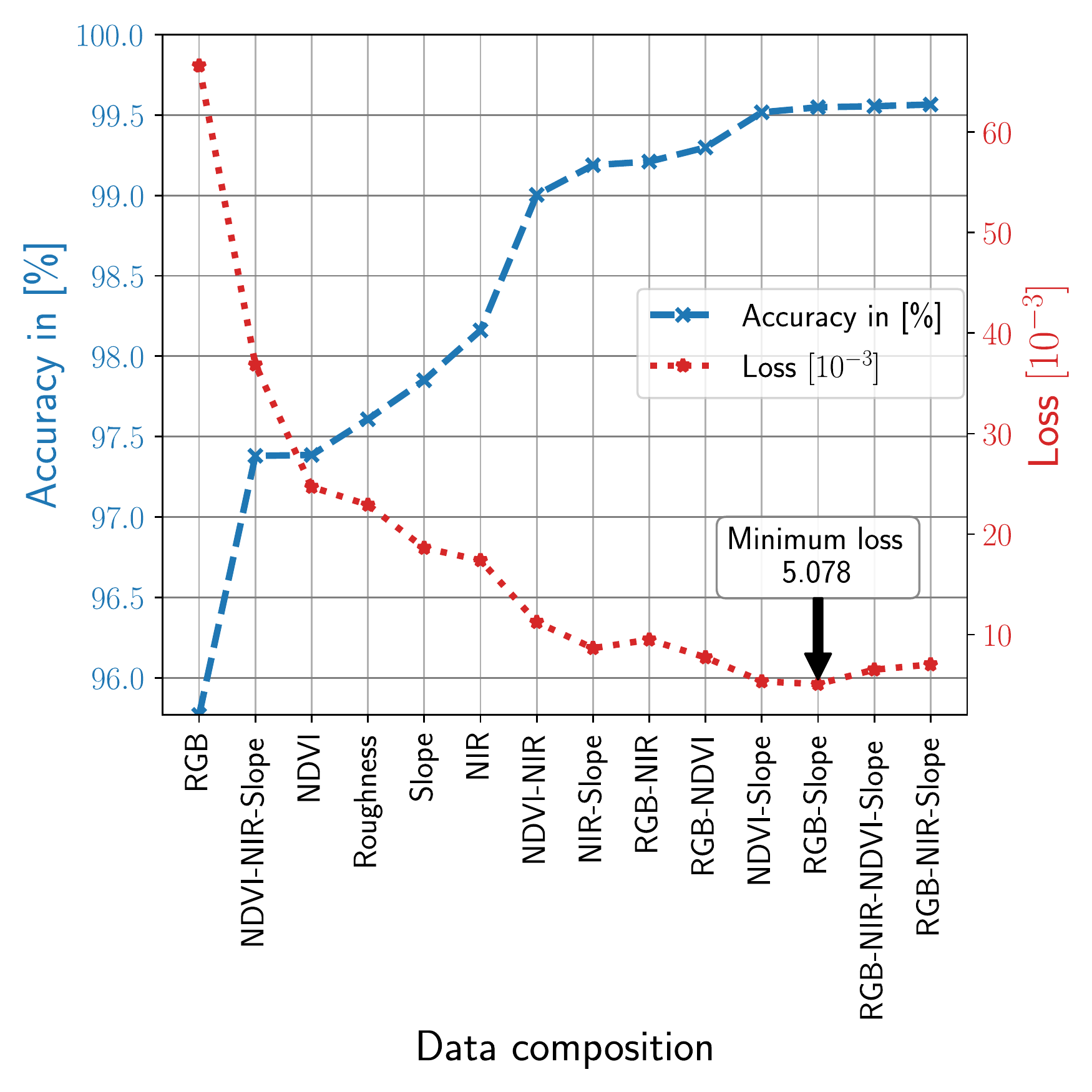}
    }
    \subfloat[Data compositions, SW 16 m$^2$]{
        \includegraphics[width=.32\textwidth]{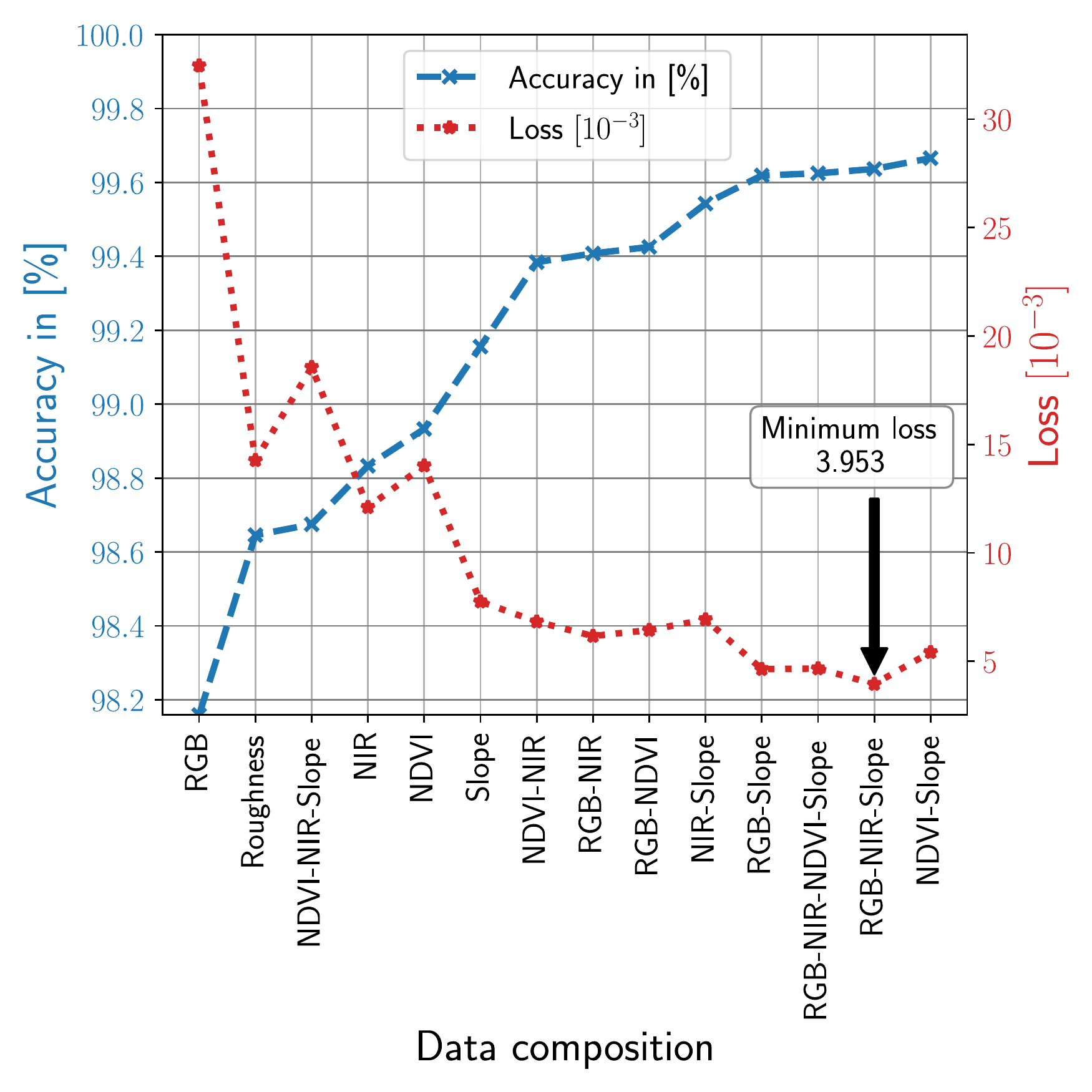}
    }
    \subfloat[Data compositions, SW 32 m$^2$]{
        \includegraphics[width=.32\textwidth]{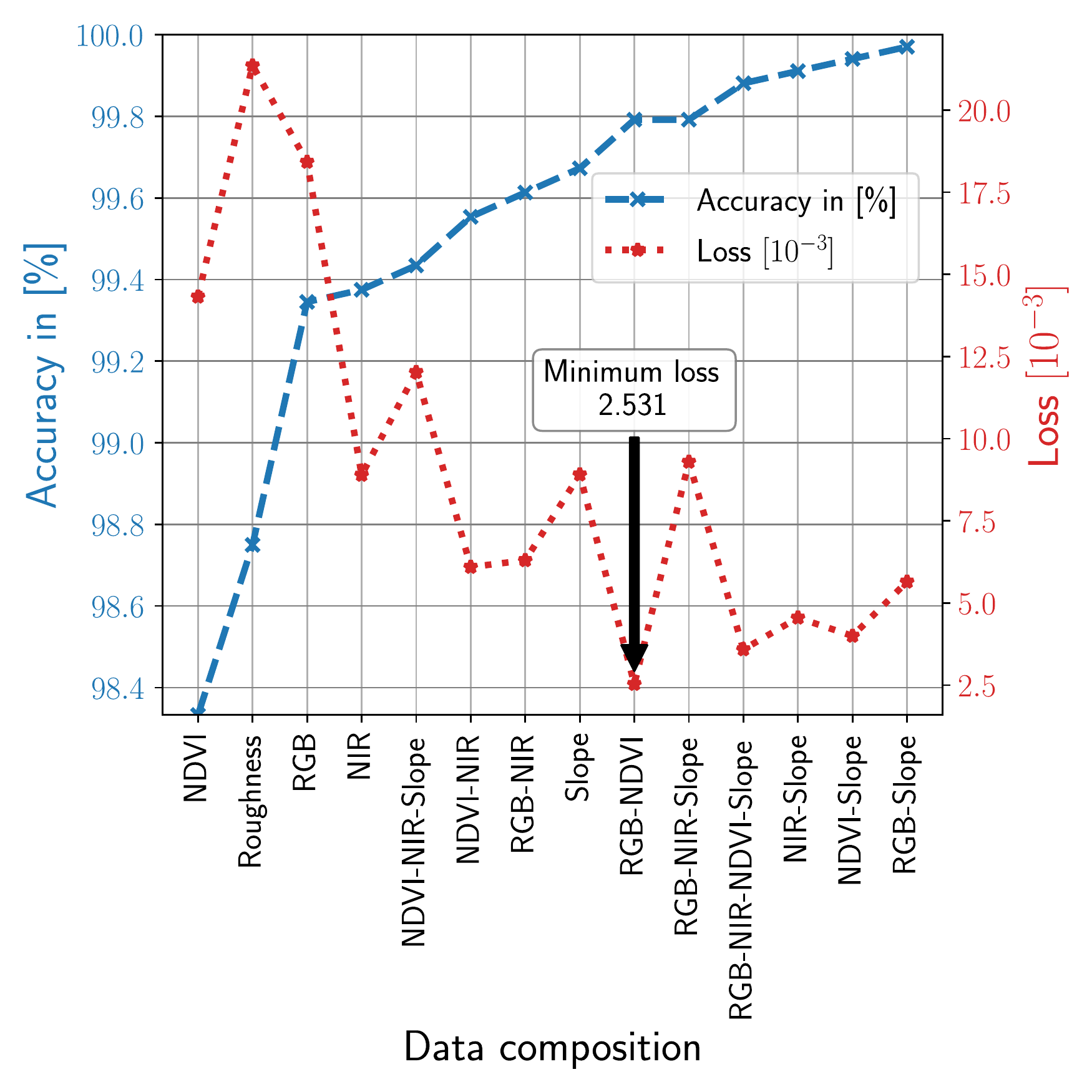}
    }\\
    \subfloat[Models, SW 8 m$^2$]{
        \includegraphics[width=.32\textwidth]{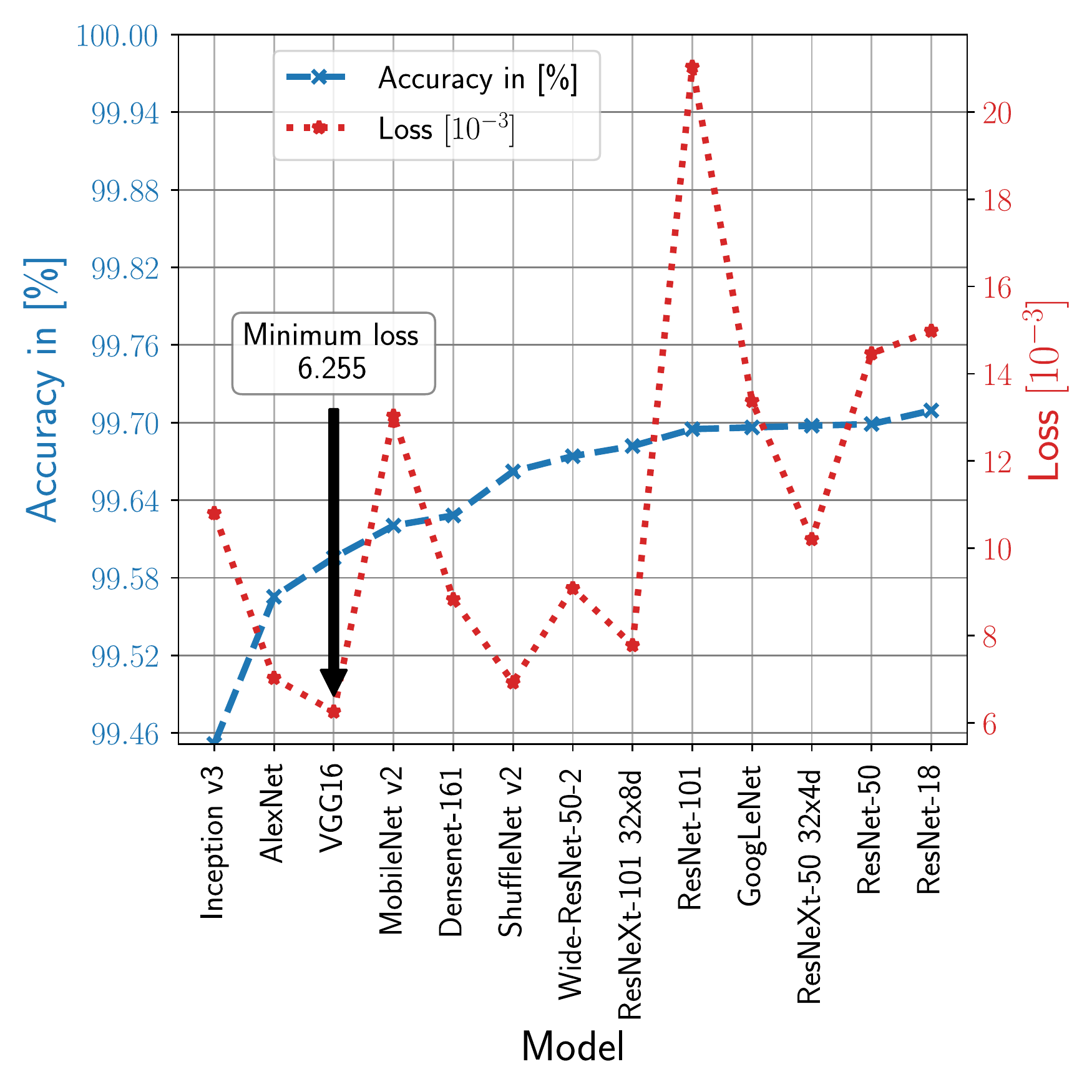}
    }
    \subfloat[Models, SW 16 m$^2$]{
        \includegraphics[width=.32\textwidth]{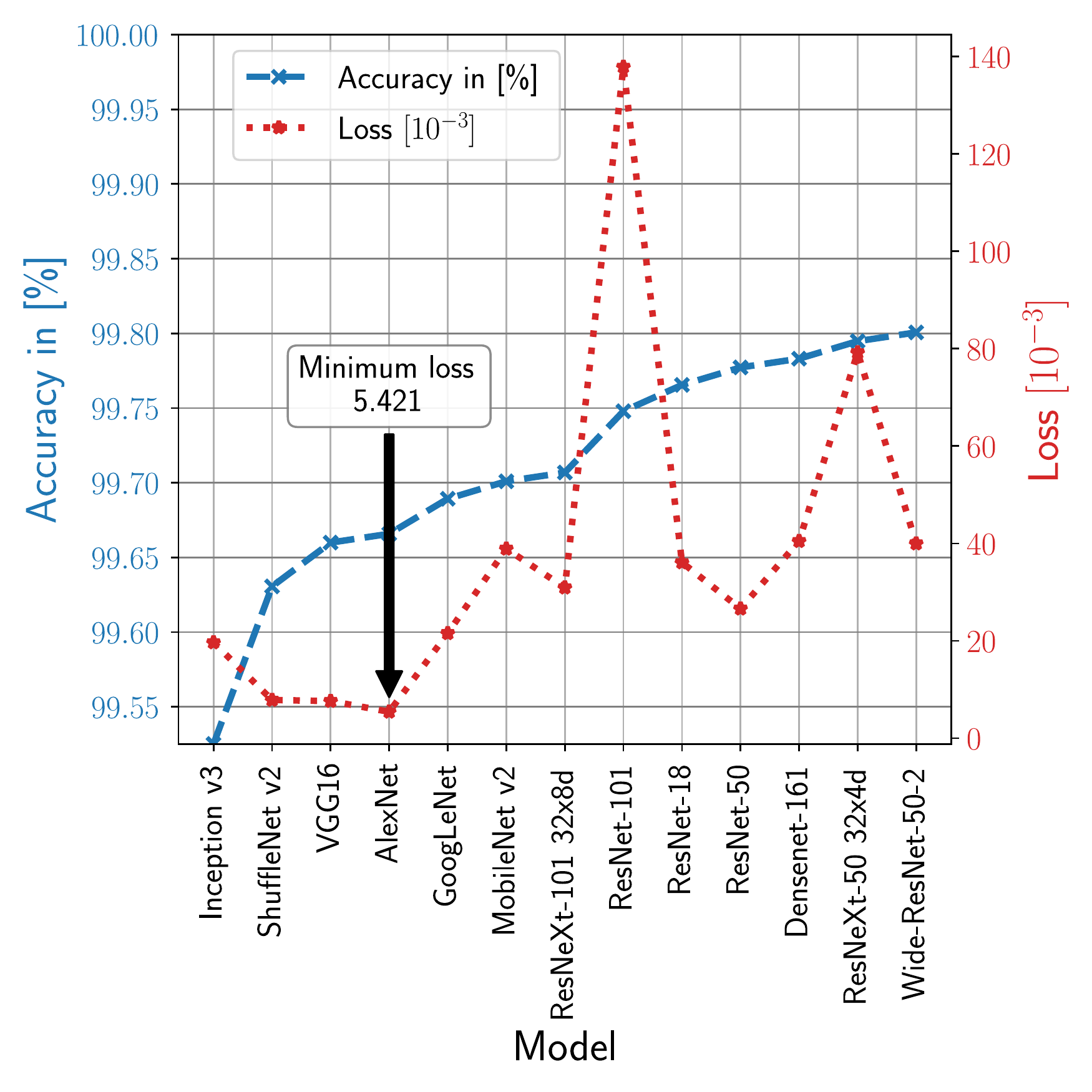}
    }
    \subfloat[Models, SW  32 m$^2$]{
        \includegraphics[width=.32\textwidth]{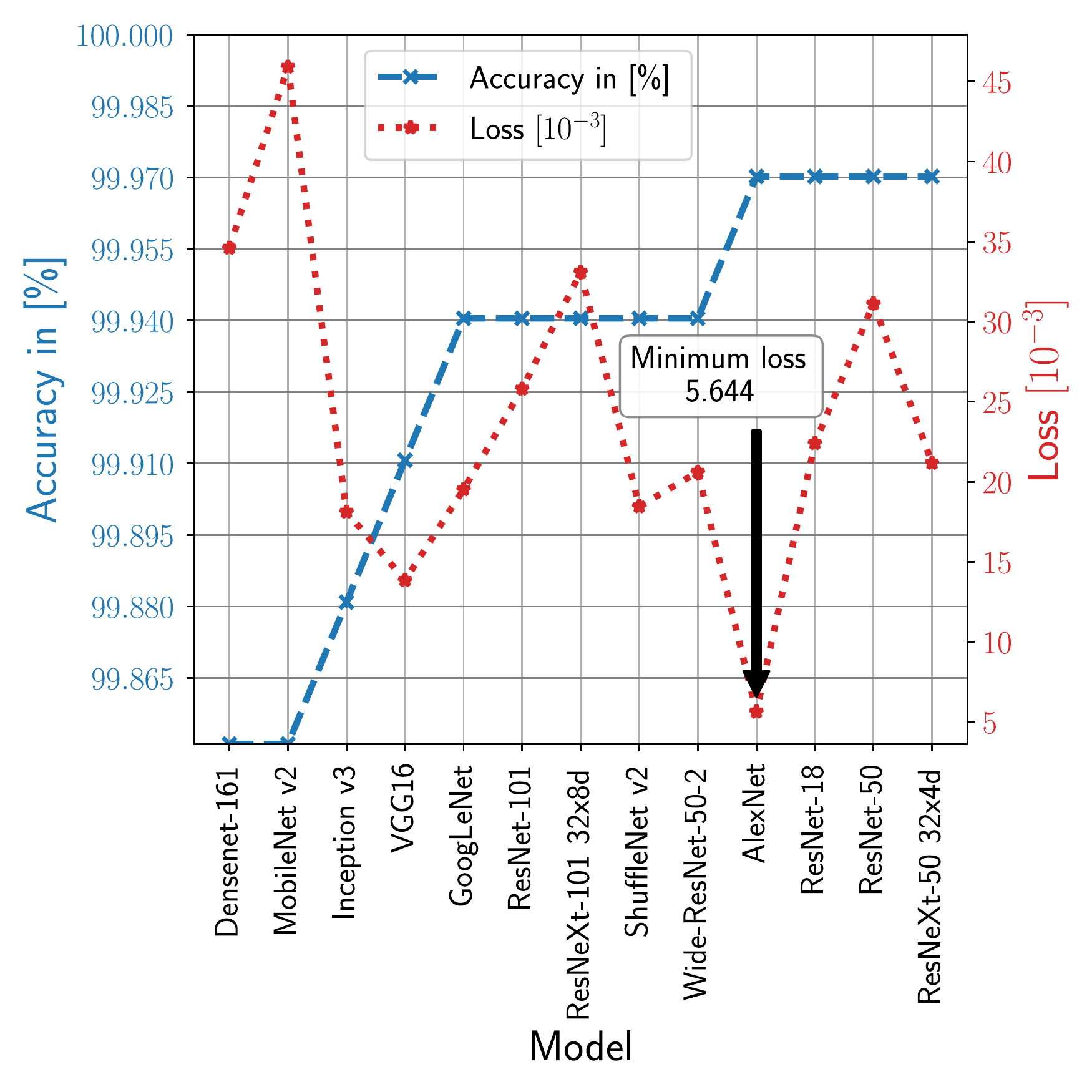}
    }
    \caption{Maximum accuracy and corresponding loss for the data compositions as well as models.}
    \label{fig:max_acc}
\end{figure*}

As proposed earlier, different SWs seem to benefit from distinct data compositions. For a SW of 8 m$^2$ the best test result is achieved on RGB-NIR-Slope data composition. About 332 test samples were classified wrong. Interestingly, the validation accuracy as well as the test accuracy reach higher values than the accuracy achieved during training. We tried to avoid this by a uniform distribution of suitable and unsuitable samples in each dataset. It might be, that the training dataset is harder to classify than the validation and test dataset. Another reason could be the usage of regularization methods like dropout during training, e.\,g. AlexNet utilizes dropout in its linear classifier layers. In Fig. \ref{fig:max_acc} the maximum accuracy and its corresponding loss values scored on test dataset are plotted.

Figure \ref{fig:max_acc} (a) shows the best results for each data composition, except for DOM which is also omitted in Fig. \ref{fig:max_acc} (b) and (c) regarding the low accuracy values scored by AlexNet. This might be caused by the normalization and the properties inherited by itself. For normalization we used the highest and lowest elevation value occurring in the dataset. If we would apply the trained model on a different area (with other minimum and maximum elevation values) the classification might be even worse. The minimum loss value is achieved by the RGB-Slope data composition. Nevertheless, the accuracy has proven better for all four runs on the RGB-NIR-Slope data composition which results also in a higher average accuracy.

For a SW of 16\,m$^2$ the NDVI-Slope data composition results in the peak accuracy of 99.665\% on the test dataset which sums up to only 57 false classifications. Furthermore, the necessary computation time is 10 times faster for training, validation and testing the neural network compared to the measured summed time achieved for the RGB-NIR-Slope data composition with a SW of 8\,m$^2$. This is caused by the lower dimensional dataset (NDVI-Slope: 2 layers, RGB-NIR-Slope: 5 layers) and the total sizes of the datasets. The dataset for the SW of 16\,m$^2$ is about four and half times smaller. Moreover, the calculated loss values are smaller in all considered phases for the NDVI-Slope data composition.  Figure \ref{fig:max_acc} (b) shows that the minimum loss is about two times smaller with the RGB-NIR-Slope data composition. However, the accuracy achieved by all four iterations on the NDVI-Slope data composition is higher. Thus, we selected the NDVI-Slope data composition in our further analysis.

\begin{figure}[ht]
    \centering
    \subfloat[DOP-RGB]{
        \includegraphics[width=.13\textwidth]{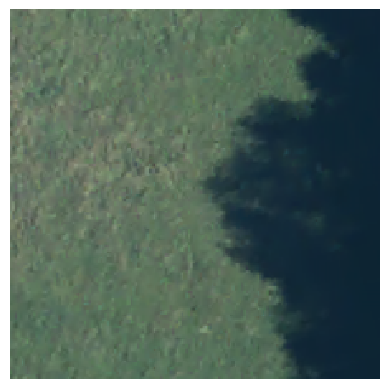}
    }
    \subfloat[NDVI]{
        \includegraphics[width=.155\textwidth]{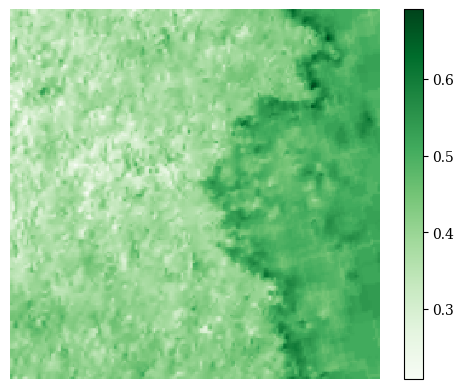}
    }
    \subfloat[Slope]{
        \includegraphics[width=.155\textwidth]{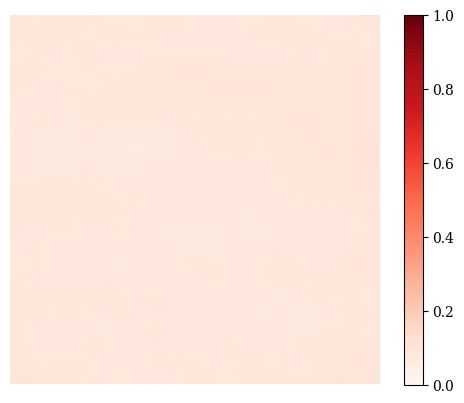}
    }\\
    \subfloat[LRP NDVI-Slope]{
        \includegraphics[width=.165\textwidth]{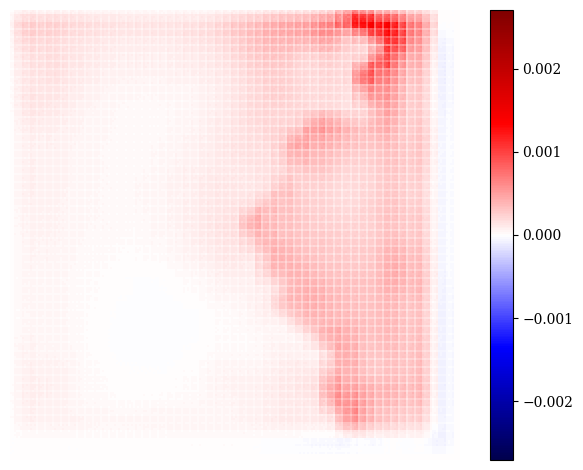}
    }
    \subfloat[LRP RGB-Slope]{
        \includegraphics[width=.165\textwidth]{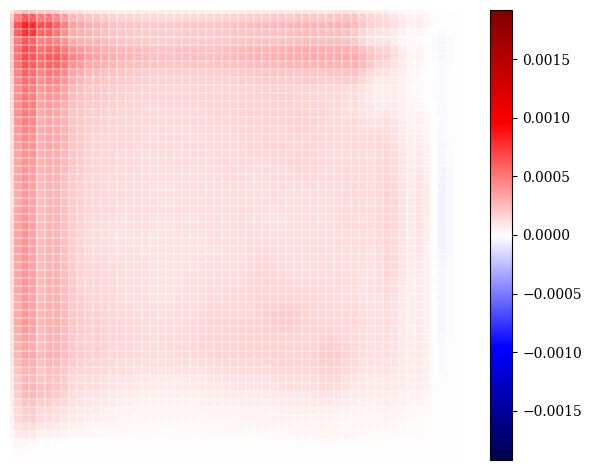}
    }
    \caption{SW 32\,m$^2$, comparison of LRP heatmaps, AlexNet trained with RGB-NDVI and RGB-Slope.}
    \label{fig:lrp}
\end{figure}

The RGB-Slope data composition with SW 32\,m$^2$ achieved a test accuracy of 99.97 \% which means, that only one example has been wrongly classified.  Figure \ref{fig:max_acc} (c) can be obtained that the RGB-NDVI data composition reached the lowest loss values (2.531 $\cdot 10^{-3}$) with a relative high accuracy. We further investigated the classification results with layer-wise relevance propagation (LRP)\cite{MBLS+2019} as shown in Fig \ref{fig:lrp}. 

The input data layers are presented in Fig. \ref{fig:lrp} (a) - (c). In Fig. \ref{fig:lrp} (d) and (e) LRP heatmaps for AlexNet trained on RGB-NDVI and RGB-Slope for an landable sample are illustrated. The darkness of red color determines how much the corresponding input pixels were relevant for the classification. If an pixel appears in blue color, these pixels contribute to the contrary prediction. AlexNet trained on RGB-NDVI predicted the sample as unsuitable for an emergency landing (false-positive) and the same model trained on RGB-Slope forecast the sample as landable (true-positive). Obviously, the model trained with RGB-NDVI data composition have issues with input samples covered by shadows. The shadows in the sample contribute with a undeniable share to the prediction (unlandable), while the models forecast trained with RGB-Slope data composition seems to be uninfluenced by the occurrence of shadows.  For that reason, we have chosen the RGB-Slope data composition for our subsequent investigations. Furthermore, both heatmaps show that especially the right input data boarder inhibit the prediction confidence. This might be caused by the applied zero padding (2$\times$2) in AlexNets architecture.

\subsection{Best transfer learning models selection}\label{subsec:model}
The selection of the appropriate model depicts a crucial task for the classification performance. The following models are considered: ResNet-18, ResNet-50, ResNet-101 \cite{HZRS2015}, AlexNet \cite{KSH2012}, VGG16 \cite{SiZi2014}, Densenet-161 \cite{HLW2016}, Inception v3 \cite{SVIS+2015}, GoogLeNet \cite{SLJS+2014}, ShuffleNet v2 with 1.0 $\times$ output channels \cite{MZZS2018}, MobileNet v2 \cite{SHZZ+2018}, ResNeXt-50 32x4d, ResNeXt-101 32x8d \cite{XGDT+2016}, Wide-ResNet-50-2 \cite{ZaKo2016}. The training, validation and testing is configured as described in Sec. \ref{subsec:dc}. The best results achieved regarding accuracy on test dataset is shown in Tab. \ref{tab:best_model}

\begin{table}[ht]\caption{The results achieved for training, validation and test of the chosen model on the corresponding input data composition. The abbreviation SW denotes the search size and T the duration time.}
    \centering\resizebox{.49\textwidth}{!}{%
    \begin{tabular}{c|c|c|c|c|c|c|c|c}
        \hline
        \multirow{1}{*}{\textbf{SW}} & \multirow{1}{*}{\textbf{Model}} & \multicolumn{3}{c|}{\textbf{Accuracy [\%]}} & \multicolumn{3}{c|}{\textbf{Loss $\pmb{[10^{-3}]}$}} & \multirow{1}{*}{\textbf{T}}\\
        $\pmb{[m^2]}$&\textbf{name}&\textbf{train}&\textbf{valid}&\textbf{test}&\textbf{train}&\textbf{valid}&\textbf{test}&\textbf{[s]}\\\hline
        8 & ResNet-18 & 99.555 & 99.726 & 99.709 & 10.612 & 14.966 & 14.980 & 702 \\\hline
        16 & Wide-ResNet-50-2 & 99.726 & 99.852 & 99.801 & 11.018 & 40.142 & 39.979 & 382\\\hline
        32& AlexNet & 99.854 & 99.917 & 99.970 & 5.735 & 5.886 & 5.644 & 19\\\hline
    \end{tabular}}
    \label{tab:best_model}
\end{table}

Obviously, the ResNet-18 model achieved the highest test accuracy with 99.709\% for the SW 32\,m$^2$. Compared to the results proposed in Tab. \ref{tab:best_res_data} for AlexNet, about 110 test samples are less incorrectly classified by ResNet-18. The loss values during investigating ResNet-18 increased compared to AlexNet and is more than twice as high as the loss value computed for VGG16 (see Fig. \ref{fig:max_acc} (d)). Nevertheless, the high accuracy during test leads to the selection of ResNet-18 for our subsequent hyperparameter optimization.

The Wide-ResNet-50-2 model reached the best accuracy of 99.801\% on the test dataset for SW 16\,m$^2$, about 34 (ca. 60\%) test samples are less wrong classified compared to the results stated in Tab. \ref{tab:best_res_data}. However, the loss achieved by Wide-ResNet-50-2 is more than seven times higer than the value achieved by AlexNet, which depicts the lowest loss as shown in Fig. \ref{fig:max_acc} (e). Due to the high margin of the accuracy between AlexNet and Wide-ResNet-50-2, the latter is chosen for the hyperparameter optimization.

For the SW 32\,m$^2$, ResNet-50, ResNet18, ResNeXt-50 32x4d and AlexNet achieved the same accuracy of 99.97\%. The calculated loss for these neural networks differ dramatically as shown in Fig. \ref{fig:max_acc} (f). The calculated loss for AlexNet is much smaller compared to all other networks. Additionally, the processing time required by AlexNet depicts the smallest demand. For that reason, AlexNet is selected for the subsequent hyperparameter optimization.

\subsection{Hyperparamter optimization}\label{subsec:ho}

The hyperparamter optimization of the selected models is performed by the utilization of \texttt{Ax}-API for the following hyperparameter: learning rate $\in [10^{-7}, 0.5]$, weight decay $\in [10^{-8}, 0.5]$, optimizer $\in$ $[$Adadelta \cite{Zei2012}, Adagrad \cite{DHS2011}, Adam, Adamax \cite{KiBa2014}, AdamW \cite{LoHu2018}, ASGD \cite{PoJu1992}, RMSprop \cite[p. 303-305]{Goo2016}, SGD \cite[p. 290-292]{Goo2016}$]$, loss function $\in [$BCELoss, MSELoss$]$. The framework offers off-the-shelve Bayesian Optimization \cite{LKOB2019} and Bandit Optimization based on Thompson sampling \cite{RVKO2017}. The objective was the optimization of the validation accuracy. The best hyperparameter configuration for each SW with the number of trials is shown in Tab. \ref{tab:ho_config}. 

\begin{table}[ht]\caption{Hyperparameter configuration identified by Bandit and Bayesian optimization.}
    \centering\resizebox{.49\textwidth}{!}{%
    \begin{tabular}{c|c|c|c|c|c|c}
        \hline
        \textbf{SW [$\bm{m^2}$]} & \textbf{Trials} & \textbf{Learning rate} & \textbf{Weight decay} & \textbf{Optimizer} & \textbf{Criterion} & \textbf{Feature extraction}\\\hline
        8 & 100 & 1.318$\cdot 10^{-2}$ & 7.661$\cdot 10^{-8}$& Adadelta & BCELoss & False\\\hline
        16 & 100 & 8.178$\cdot 10^{-6}$ & 3.140$\cdot 10^{-4}$ & AdamW & BCELoss & False\\\hline
        32 & 500 & 3.087$\cdot 10^{-5}$ & 2.115$\cdot 10^{-2}$ & AdamW & MSELoss & False\\\hline
    \end{tabular}}
    \label{tab:ho_config}
\end{table}

Obliviously, adjusting all weights results in the highest training and validation accuracy values during the hyperparameter optimization. The achieved results are reported in Tab. \ref{tab:best_ho_results}. 

\begin{table}[ht]\caption{Results achieved by the hyperparameter optimization.}
    \centering\resizebox{.49\textwidth}{!}{%
    \begin{tabular}{c|c|c|c|c|c|c|c|c|c|c}
        \hline
        \textbf{SW} & \multicolumn{3}{c|}{\textbf{Accuracy [\%]}} & \multicolumn{3}{c|}{\textbf{Loss $\pmb{[10^{-3}]}$}} & \multicolumn{3}{c|}{\textbf{Precision [\%]}}&\textbf{T}\\
        $\pmb{[m^2]}$&\textbf{train}&\textbf{valid}&\textbf{test}&\textbf{train}&\textbf{valid}&\textbf{test}&\textbf{train}&\textbf{valid}&\textbf{test}&\textbf{[s]}\\\hline
        8 & 99.998 & 99.965 & 99.958 & 0.135 & 1.973 & 1.691 & 99.999 & 99.967 & 99.984 & 524 \\\hline
        16 & 100 & 99.959 & 99.959 & 0.027 & 1.023 & 1.334 & 100 & 99.967 & 99.976 & 764 \\\hline
        32 & 99.937 & 99.958 & 99.940 & 0.626 & 0.592 & 0.508 & 100 & 100 & 99.940 & 17 \\\hline
    \end{tabular}}
    \label{tab:best_ho_results}
\end{table}

\begin{figure*}[ht]
    \centering
    \subfloat[SW 32\,m$^2$, Coordinates: 428849.6, 5696761.4]{
        \includegraphics[width=.235\textwidth]{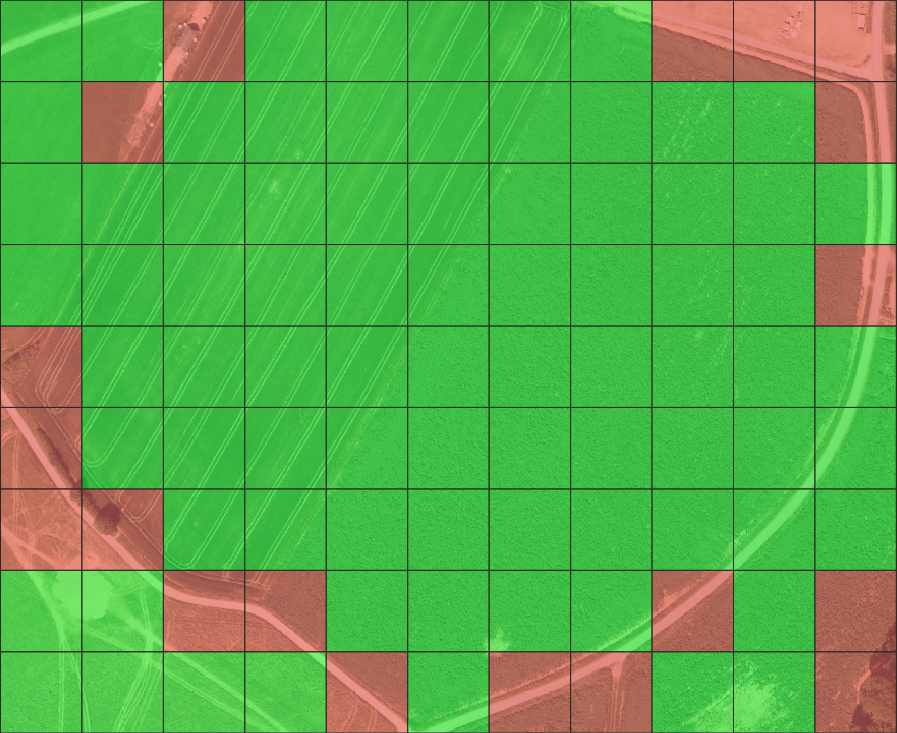}
    }
    \subfloat[Ensemble transfer learning]{
        \includegraphics[width=.235\textwidth]{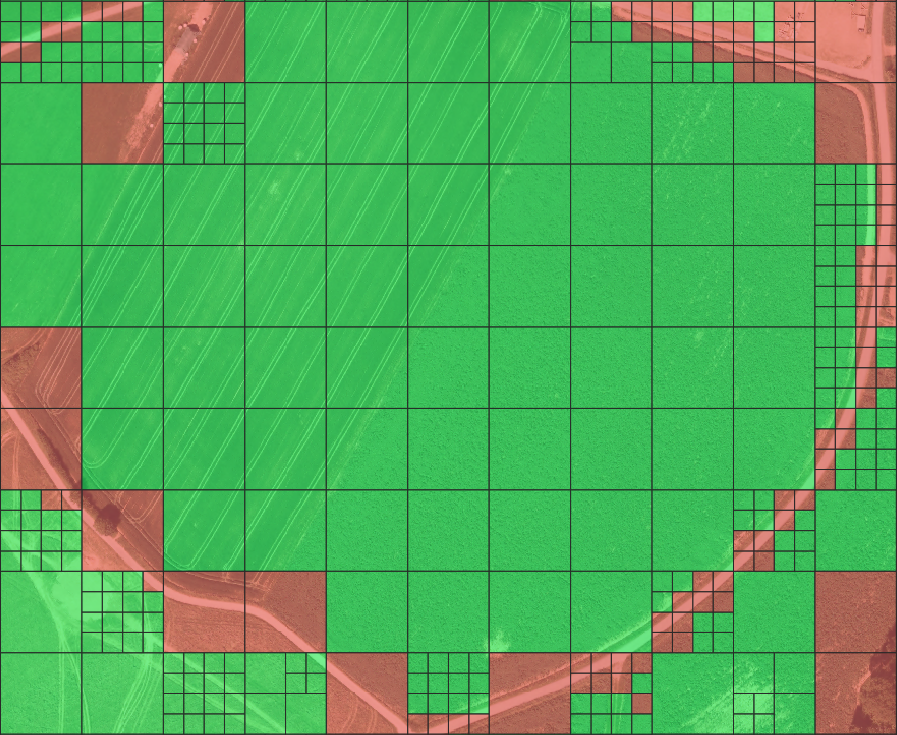}
    }
    \subfloat[DOP, difficult area, Coordinates: 426518.1, 5696684]{
        \includegraphics[width=.28\textwidth]{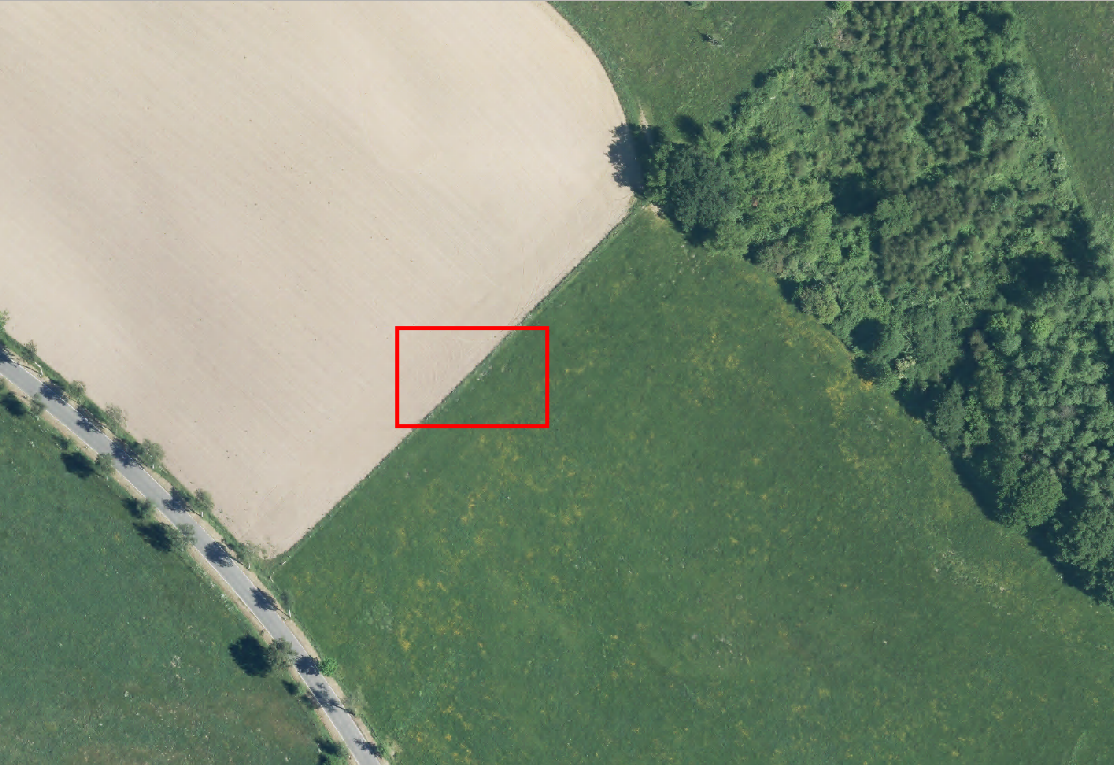}
    }\\
    \subfloat[Hillshade of a ditch]{
        \includegraphics[width=.25\textwidth]{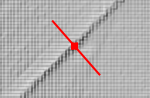}
    }
    \subfloat[Slope profile in \% over distance]{
        \includegraphics[width=.50\textwidth]{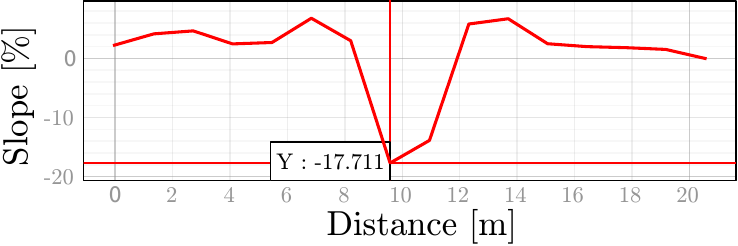}
    }\\
    \subfloat[Hillshade, ensemble transfer learning]{
        \includegraphics[width=.28\textwidth]{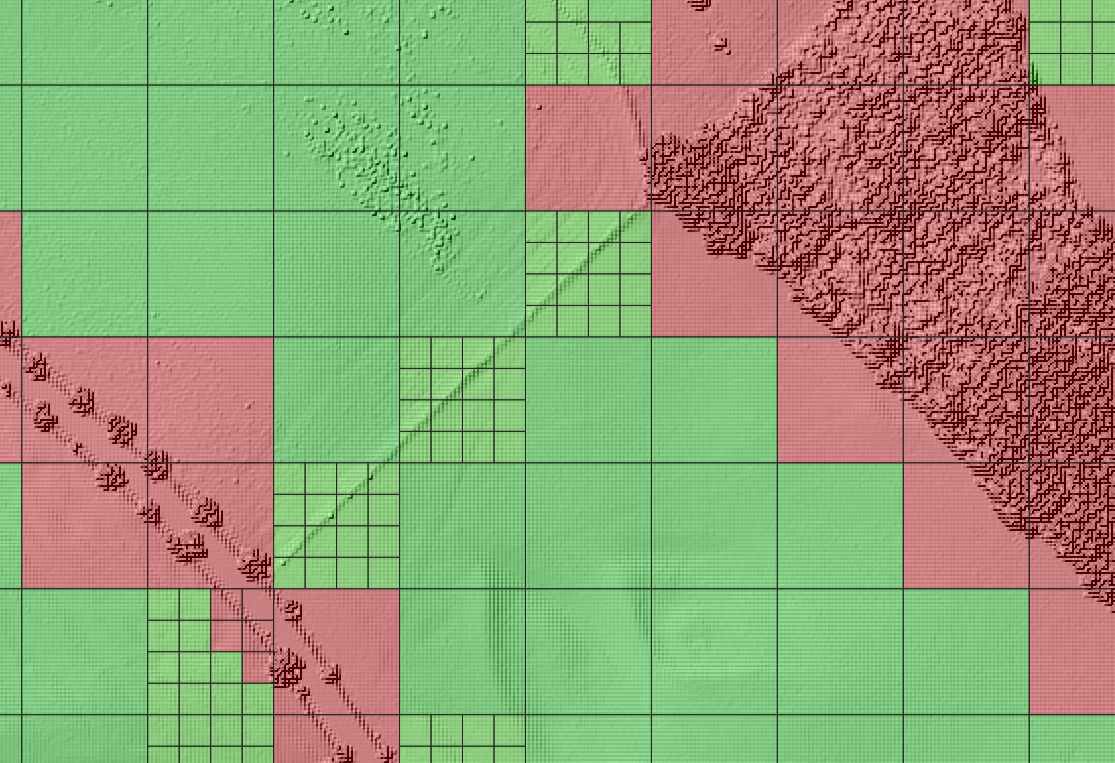}
    }
    \subfloat[Hillshade, enhanced model]{
        \includegraphics[width=.28\textwidth]{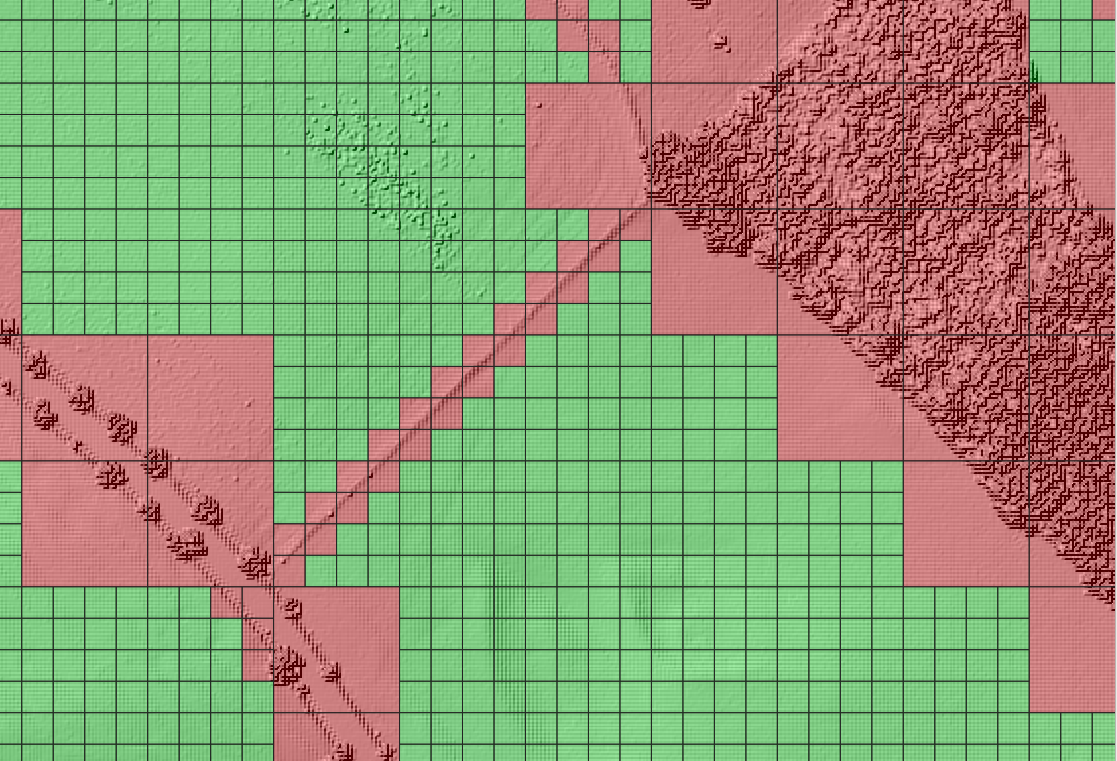}
    }
    \subfloat[Hillshade, identified ELFs]{
        \includegraphics[width=.285\textwidth]{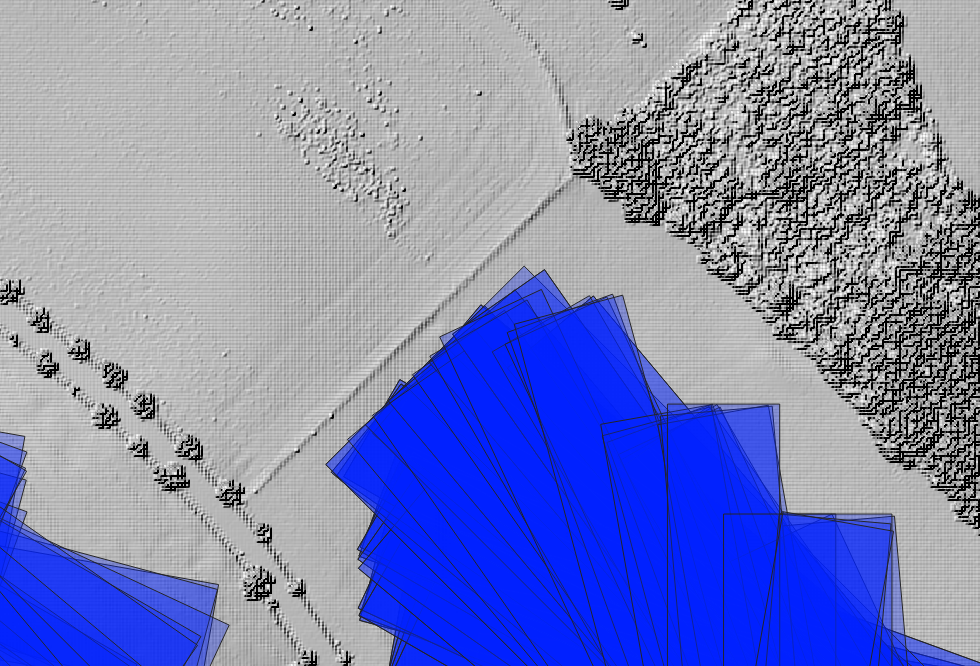}
    }
    \caption{Ensemble transfer learning model segmentation and ELF identification.}
    \label{fig:rws}
\end{figure*}

The hyperparameter optimization is composed of 100 trials for SW 8\,m$^2$ and 16\,m$^2$ and due to the low computational demand 500 trials of hyperparameter optimization are conducted for SW 32\,m$^2$. The results show, that the accuracy has increased dramatically for SW 8\,m$^2$ and 16\,m$^2$ during the hyperparameter optimization. Considering the accuracy reached during test for SW 8\,m$^2$, the number of false classification has shrunken to 32 examples. Compared to the results achieved by ResNet-18 reported in Tab. \ref{tab:best_model} the number of false classified examples is about 7 times smaller. The loss values presented in Tab. \ref{tab:best_ho_results} are more than 7 times reduced against the values stated in Tab. \ref{tab:best_model}. Additionally, we consider the precision for monitoring the reliability of our classification as suitable for an emergency landing. Fortunately, the precision during training, validation and test is close to 100\% for each SW. The accuracy improvement regarding SW 16\,m$^2$ is quite high. On the test dataset the model classified about 7 samples wrong. Besides, the loss during training, validation and test dropped at least about 30 times compared to the values reported in Tab. \ref{tab:best_model}. The accuracy values for SW 32\,m$^2$ are comparable to the results in Tab. \ref{tab:best_model}, however, the loss values decreased at least 9 times against the presented once in Tab. \ref{tab:best_model}. The newly trained models with the best identified hyperparamter configuration are used for facilitating the subsequent ensemble learning.

\subsection{Ensemble transfer learning model and runway identification}

The proposed ensemble transfer learning model is build up in an hierarchical matter. It is composed of the best performing input data composition (Sec. \ref{subsec:dc}), the corresponding model (Sec. \ref{subsec:model}) and its hyperparameters (Sec. \ref{subsec:ho}) for each SW. The model consists of three siblings which are already powerful by themselves. The predictions of each model are weighted by their confidence. The confidence is calculated as follows: $\frac{\texttt{max(softmax(x))}-0.5}{0.5}$. The inference of the model trained for SW 32\,m$^2$ is performed on the whole area of about 247.581\,km$^2$ e.\,g. see Fig. \ref{fig:rws} (a). 

Afterwards, the regions are selected with a lower prediction confidence than 99\% or classification as landable -- to increase the reliability of landable classifications. For these areas the inference of the model trained for SW 16\,m$^2$ is conducted. Thus, the area of interest for the inference shrinks to $\approx 30.328$\,km$^2$ so that the computational demand is reduced more than 8 times. Subsequently, the model trained for SW 8\,m$^2$ is applied for areas with an averaged prediction confidence $<99$\% or a voted classification as landable. The resulting voted prediction is presented in Fig \ref{fig:rws} (b). Obviously, the road is better identified by our hierarchical ensemble transfer learning model, compared to the segmentation performed by one single model. We were able to segment 247.581\,km$^2$ in 27.143\,km$^2$ as landable and 220.438\,km$^2$ as unlandable. 

It became obvious, that the ensemble transfer learning model still lacks in some problematic areas like pasture fence or ditches between fields, a sample of the latter is shown in Fig. \ref{fig:rws} (c) and (d). Figure \ref{fig:rws} (c) presents the DOP of the ditch between two fields with a red rectangle which tags the area illustrated in Fig. \ref{fig:rws} (d) as hillshade transformation of the DSM data. The red line in Fig. \ref{fig:rws} (d) covers the points used for the slope profile in Fig. \ref{fig:rws} (e) and the red point tags the maximum negative slope position of the slope profile. The slope discovers loss till -17.711\%, which clearly indicates that emergency landing would be quite dangerous there. Nevertheless, in Fig. \ref{fig:rws} (f) the proposed ensemble transfer learning model clearly classified the corresponding patches as landable. For that reason, we created a dataset for SW 8\,m$^2$ which is composed as follows: \{train: 12,348 with \{0:6,173, 1:6,175\}, test: 2059 with \{0:1,029, 1:1,030\}\}. The classification has been performed by hand with the main focus on the exclusion on the afore mentioned difficult areas.

The already well proven ResNet-18 architecture has been selected. The models hyperparameters became improved as before with \texttt{Ax}-API and the same search space. The optimization lead to the hyperparameters shown in Tab. \ref{tab:ho_config_2}. 

\begin{table}[ht]\caption{Hyperparameter config., Resnet-18, SW 8\,$m^2$.}
    \centering\resizebox{.49\textwidth}{!}{%
    \begin{tabular}{c|c|c|c|c|c|c}
        \hline
        \textbf{SW [$\bm{m^2}$]} & \textbf{Trials} & \textbf{Learning rate} & \textbf{Weight decay} & \textbf{Optimizer} & \textbf{Criterion} & \textbf{Feature extraction}\\\hline
        8 & 100 & 1.653$\cdot 10^{-2}$ & 4.607$\cdot 10^{-7}$& ASGD & BCELoss & False\\\hline
        \end{tabular}}
    \label{tab:ho_config_2}
\end{table}

Afterwards, the new model has been applied, where the average confidence was less than 99\% and the voted classification equals landable. As a consequence, the model has been applied for about 7.562\,$m^2$. The final patch segmentation for the same problematic area mentioned before is shown in Fig. \ref{fig:rws} (g). Evidently, the ditch has been correct classified as not suitable for an emergency landing. The analyzed area has been subdivided into 26.252\,$m^2$ as landable and 221.329\,$m^2$ as not suitable for an emergency landing.

Subsequently, the ELF search in the landable areas has been performed for an inclination of 18.66\% -- inherited from airport Courchevel. Together with this extreme, initial uphill slope assumption, Eq. \ref{eq:sgfinal} and Alg. \ref{alg:rws} -- \texttt{rotation\_angles}$\in [0,4,...,179]$ and \texttt{stride} $\frac{width}{2}$ -- in sum 115,188 ELFs have been identified with a minimum length of 151.877\,m.

\begin{algorithm}[ht]\caption{Identifying rectangular shaped ELFs.}\label{alg:rws}
\scriptsize
    \DontPrintSemicolon
    \SetKwInput{KwInput}{Input}                
    \SetKwInput{KwOutput}{Output}              
    \SetKwFunction{getCentroid}{get\_centroid}
    \SetKwFunction{getElf}{get\_elf}
    \SetKwFunction{rotate}{rotate}
    \SetKwFunction{checkDimensions}{check\_dimensions}
    \SetKwFunction{getPolygonLimits}{get\_polygon\_limits}
    \SetKwFunction{shift}{shift}
    \SetKwFunction{contains}{contains}
    \SetKwFunction{optimizeLength}{optimize\_length}
    \SetKwFunction{outputRow}{output\_row}

    \KwInput{$polygon := \text{Georeferenced polygon},  elf\_length := float,  elf\_width := float $}
    \KwOutput{$\text{Set of rows}$}
    $rotation\_angles \gets [i\cdot \frac{\pi}{180}], \text{where }i \in [0, 4, \cdots, 179]$\\
    $centroid \gets \getCentroid(polygon)$\\
    $elf \gets \getElf(elf\_length,  elf\_width)$\\
    $stride \gets \frac{ elf\_width}{2}$\\
    \ForEach{$rotation\_angle \in rotation\_angles$}{
        \If{$rotation\_angle != 0$}{
            $poi \gets \rotate(polygon, rotation\_angle, centroid)$
        }
        \Else{
            $poi \gets polygon$
        }
        \If{$not\text{ }\checkDimensions(poi)$}
        {
            $continue$
        }
        $y\_min, y\_max, x\_min, x\_max \gets \getPolygonLimits(poi)$\\
        $\Delta y \gets (y\_max - y\_min)$\\
        $y\_start\_positions \gets (i\cdot stride), \text{where }i \in [0, 1, \cdots, \frac{\Delta y}{stride}+1]$\\
        \ForEach{y\_start\_position $\in$ y\_start\_positions}{
            $x\_shift \gets 0.0$\\
            \While{$x\_max-x\_shift >= elf\_length$}{
                $shifted\_elf \gets \shift(elf, x\_shift, y\_start\_position)$\\
                \If{$\contains(poi,shifted\_elf)$}{
                    $resize \gets 1$\\
                    \While{$\contains(poi,shifted\_elf)$}{
                        $shifted\_elf \gets \optimizeLength(shifted\_elf,resize)$\\
                        $resize \gets resize+1$
                    }
                    $x\_shift \gets x\_shift+resize+1$\\
                    $\outputRow(\rotate(shifted\_elf,-rotation\_angle,$ $centroid), resize-1)$
                }
                \Else{
                    $x\_shift \gets x\_shift + 1$
                }
            }
        }
    }
\end{algorithm}

In fact, the high number of identified ELFs underlay the benevolent assumption of 18.66\% uphill slope and therefore the reduced min. required ELF length \footnote{The number of identified ELFs can be heavily influenced by the analyzed \texttt{rotation\_angles} and \texttt{stride} during the search.}. The ELFs were further investigated regarding their appropriateness for an emergency landing with respect to the required length considering the slope. Each ELF length is investigated with Eq. \ref{eq:sgfinal}, if the minimum required dimension is fulfilled and the slope is greater than -10\%. During the investigation 54,997 ELFs still remained in our database. The identified ELFs in the problematic area are shown in Fig. \ref{fig:rws} (h).

Thereby, the maximum up- and downhill slope is 25.384\% and -9.999\% , the average up- and downhill slope is 6.127\% $\pm$ 4.505\% and -2.827\% $\pm$ 2.448\%. The maximum and minimum runway length is 893.877\,m and 152.877\,m. In \cite{EASA2014} the EASA recommends to increase the runway by a factor of 1.6 if the surface is covered by very short, wet grass with a firm subsoil. This requirement is fulfilled by 11,960 ELFs. In cases of doubt, the use of the wet factor 1.15 is recommended which is satisfied by 37,759 ELFs.

%% file: chapter4/chapter4.tex
\section{Conclusion and future works}
We proposed the generated training and test datasets as well as the deployed deeplearning infrastructure. Furthermore, we performed an in-depth analysis regrading the best input dataset configuration with AlexNet. For distinct SW different dataset configuration have proven the best results. Subsequently, we investigated the best models for the corresponding dataset configuration. The hyperparameters of each selected model are highly improved. The trained models are applied as ensemble transfer learning model to a area of 247.581\,km$^2$. Hence, we were able to identify 54,997 ELFs and saved the results in a database. 

In future works we will develop our own neural network architectures, verify the obstacle clearness of the final approach and build a recommendation system for the best suitable ELFs.